\documentclass[journal]{IEEEtran}

\usepackage{graphicx}
\usepackage{amsmath}
\usepackage{algorithm}
\usepackage{algorithmicx}
\usepackage{algpseudocode}
\usepackage{multirow}
\usepackage[table,xcdraw]{xcolor}
\usepackage{booktabs}
\usepackage{subfigure}

\newcommand{\mygreencolor}{\color{black}}
\newcommand{\mybluecolor}{\color{black}}

\ifCLASSINFOpdf
\else
\fi
\begin{document}
%
\title{Online Alternate Generator against Adversarial Attacks}
%
%
%

\author{Haofeng~Li,
        Yirui~Zeng,
        Guanbin~Li,
        Liang Lin,
        Yizhou~Yu

\thanks{This work was supported in part by the National Key Research \& Development Program (No.2020YFC2003902), in part by the Guangdong Basic and Applied Basic Research Foundation~(No.2020B1515020048), in part by the National Natural Science Foundation of China~(No.61976250, No.61702565, No.U1811463), in part by the Fundamental Research Funds for the Central Universities~(No.18lgpy63), and was also sponsored by CCF-Tencent Open Research Fund (Corresponding author is Guanbin Li).}

\thanks{H. Li is with Shenzhen Research Institute of Big Data, The Chinese University of Hong Kong (Shenzhen), 
	Shenzhen 518172, China (e-mail: lhaof@foxmail.com).}
\thanks{Y. Zeng, G. Li and L. Lin are with the school of Data and Computer Science, Sun Yat-sen University, Guangzhou 510006, China (e-mail: zengyr5@mail2.sysu.edu.cn; liguanbin@mail.sysu.edu.cn; linliang@ieee.org).}
\thanks{Y. Yu is with Deepwise AI Lab (e-mail: yizhouy@acm.org).}}

%
%

\markboth{Journal of \LaTeX\ Class Files,~Vol.~14, No.~8, August~2015}%
{Shell \MakeLowercase{\textit{et al.}}: Bare Demo of IEEEtran.cls for IEEE Journals}
%



\maketitle

\begin{abstract}
  The field of computer vision has witnessed phenomenal progress in recent years partially due to the development of deep convolutional neural networks. However, deep learning models are notoriously sensitive to adversarial examples which are synthesized by adding quasi-perceptible noises on real images. Some existing defense methods require to re-train attacked target networks and augment the train set via known adversarial attacks, which is inefficient and might be unpromising with unknown attack types. 
  To overcome the above issues, we propose a \textit{portable} defense method, online alternate generator, which does not need to access or modify the parameters of the target networks. The proposed method works by online synthesizing another image from scratch for an input image, instead of removing or destroying adversarial noises. To avoid pretrained parameters exploited by attackers, we alternately update the generator and the synthesized image at the inference stage. Experimental results demonstrate that the proposed defensive scheme and method outperforms a series of state-of-the-art defending models against gray-box adversarial attacks.
\end{abstract}


%
\IEEEpeerreviewmaketitle

\section{Introduction}
\begin{figure}[t]
	\begin{center}
		\includegraphics[width=0.8\linewidth]{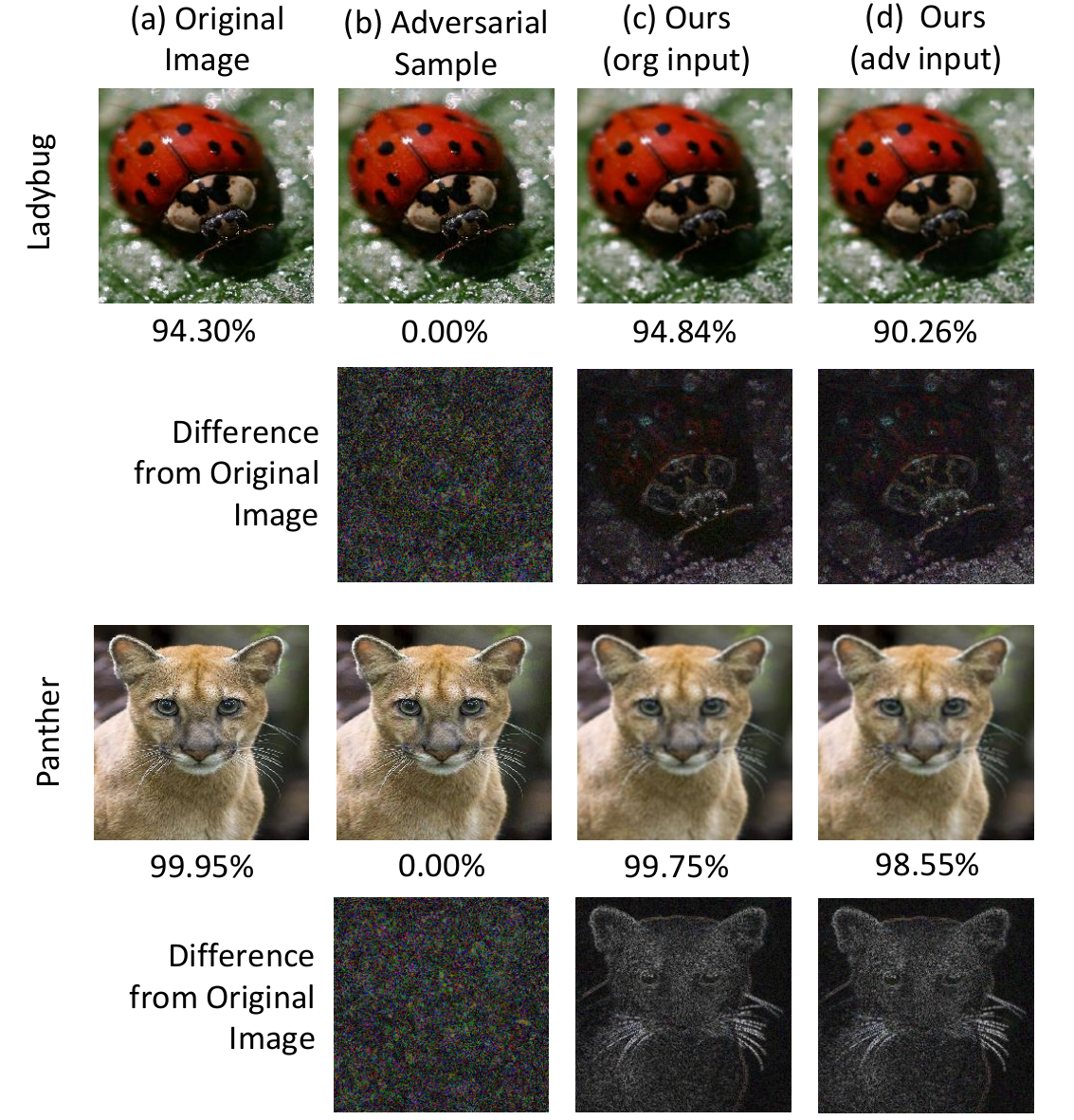}
	\end{center}
	\caption{The effectiveness of removing adversarial noise by our proposed method. (a) and (b) denote original images and their adversarial examples. (c) and (d) denote our generated results by taking original images and adversarial examples as reference respectively. The second row and the fourth row are normalized absolute difference between original images (a) and (b-d). Our results have different noise distribution from input adversarial examples. The predicted probability of correct labels is attached under each sample of (a-d). Our proposed defense method significantly increases the probability of predicting correct labels.\iffalse It is worthy notice that the results (c) obtained by taking original images as reference and the results (d) obtained by using adversarial example have similar noise distributions. Their probability of predicting correct labels are also close.\fi}
	\label{fig:simpleComp}
\end{figure}

%
%
%
%
\IEEEPARstart{I}{n} recent years, deep convolutional neural networks have obtained state-of-the-art performances on many machine learning benchmarks, since they can harvest adaptive features on large-scale training sets, in comparison to traditional methods based on handcrafted features. However, deep learning models are found to be vulnerable with \textit{adversarial attacks}, which aim at synthesizing \textit{adversarial samples} that are perceptually similar to real images but can mislead attacked models to yield totally incorrect labels, as shown in Fig.~\ref{fig:simpleComp}(b). Adversarial examples can be generated by applying quasi-perceptible perturbations which does not change labels recognized by human subjects. Such perturbations can be computed via constrained optimization or backward propagation with an incorrect supervision. Thus, given a pretrained \textit{target network} that might be accessed and attacked by hackers, how to protect it from adversarial attacks remains an important problem.

Many defense methods are developed to resist adversarial attacks on deep learning models. These defense methods can be roughly categorized into two groups. One group of defenses act as a preprocessing component which does not require accessing, modifying or re-training the attacked target network. These methods are portable and practical with different target networks or tasks since the knowledge of target networks may be confidential in real applications. These methods usually resort to image denoising and smoothing to remove adversarial noises, or image transformations that could destroy adversarial noises to some extent. Another group of defense methods require to access or re-train the parameters of target network. We argue that these methods may be impractical and inefficient in real applications. For examples, adversarial training methods need to obtain the knowledge of adversarial attacks and might be unpromising to resist \textit{unseen} attack types. Kurakin et al.~\cite{kurakin2017adversarial} also suggest that adversarial training with single-step attacks does not confer robustness to iterative adversarial samples. Ensemble adversarial training~\cite{tramer2018ensemble} requires augmenting the train set by $N \times M$ times, designing and training $N$ different target networks, which is inefficient. $M$ is the number of different known adversarial attacks used in adversarial training. It is difficult to transfer DefenseGAN~\cite{samangouei2018defense-gan} on large images since training GAN with large images is unstable and might need to adjust the network architecture for different datasets.

Motivated by the above observations, in this paper we aim at addressing such a problem: developing a \textit{portable} defense method that protects a pretrained target network from $unseen$ adversarial attacks with \textit{images of large size}.

We define a portable defense as a method that does not need to access, modify or re-train the attacked target network. We claim that developing a portable defense is important because some parameters of the target network might be commercially confidential. Re-designing and re-training the target network could cause heavy load. A portable defense allows itself as a reusable self-contained component invoked via API. Developing a defense method against unseen attacks is critical since it is impractical to know the attack type used by attackers. 
On the other hand, developing a defense method working with large images, whose size is not smaller than $200 \times 200$, is practical and meaningful. Given the same $L_\infty$ norm upper bound, the number of possible adversarial samples increases exponentially with the number of pixels. Thus defending against adversarial attacks with larger images is more difficult. Tramer et al.~\cite{tramer2018ensemble} also suggest that results obtained on simple datasets~\cite{lecun1998gradient} with small images does not always generalize to harder tasks, for example, a classification benchmark with larger images.

To address the above-mentioned issues, we conceive a novel defending framework, online alternate generator. The proposed defense scheme works by online synthesizing an image that shares the same semantics with the input image but is almost free from adversarial noises. 
To avoid the model parameters stolen and exploited by attackers, we also propose to update the generator and the synthesized image at the inference stage, in an iterative and alternate manner. Besides, Gaussian noise is utilized as an additional perturbation in updating the synthesized image, to prevent the generator from fitting adversarial noises.

Our proposed method enjoys the following strengths. First, since the proposed method does not require any knowledge of target networks or adversarial attacks, it is a \textit{portable} defense that can theoretically protect arbitrary target classifiers from arbitrary \textit{unseen} adversarial attacks. Second, a Gaussian perturbation is added to yield an image, which not only introduces randomness but also decreases the probability of fitting adversarial noise on the given input. Third, as the proposed method adopts online training, its model parameters are not fixed during inference and cannot be accessed beforehand by potential attackers in real scene to synthesize an adversarial example.

In summary, this paper has the following contributions:
\begin{itemize}
	\item We develop a novel \textit{portable} defense framework, online alternate generator, which can resist unseen adversarial attacks for unlimited pretrained classifiers, without the knowledge of target networks or adversarial attacks.
	\item We propose a stopping criteria which does not require accessing any adversarial samples such that the proposed method can deal with unseen attacks.
	\item We verify the transferability and generalization of the proposed method with different adversarial attacks, target models and benchmarks. Different selections of hyper-parameters are also well investigated.
	\item This paper presents extensive experimental results to verify that the proposed framework surpasses a wide range of existing state-of-the-art defense algorithms.
\end{itemize}

\section{Related Work}
\subsection{Adversarial Attack}
Deep convolutional neural networks have demonstrated powerful fitting capacity in solving computer vision problems for last several years, but they are threatened by adversarial attacks~\cite{szegedy2014intriguing,goodfellow2015explaining,moosavidezfooli2016deepfool,xie2017adversarial,metzen2017universal,moosavidezfooli2017universal,Houdini2017,dong2018boosting,poursaeed2018generative,zhao2018generating,arnab2018on,xiao2018spatially}.
Fast Gradient Sign Method (FGSM)~\cite{goodfellow2015explaining} synthesizes adversarial examples by adding some weighted gradients which increases the prediction loss of the attacked target network, as shown in $I_{adv} = I + \epsilon \cdot sign(\partial L(F(I,\theta),Y_I) / \partial I)$
where $I$ denotes a real image and $I_{adv}$ is the generated adversarial example. $sign(\cdot)$ returns 1 for positive input, and returns -1 for negative one. $L(x,y)$ denotes a loss function that estimates the difference between a prediction $x$ and the ground truth $y$. $F(\cdot,\theta)$ is the attacked target neural network with parameters $\theta$. $Y_I$ denotes the ground truth annotation of image $I$. $\epsilon$ is the $l_\infty$ norm of the adversarial perturbation and the weight of the gradient with respect to $I$. Attack strength can be controlled by $\epsilon$.
The above algorithm is referred as \textit{untargeted} attack. A \textit{targeted} variant~\cite{kurakin2017adversarial} of FGSM encourages the attacked network to predict high probability at some deliberately incorrect category $Y_{adv}$,
as formulated in $I_{adv} = I - \epsilon \cdot sign\left(\partial L\left(F\left(I,\theta\right),Y_{adv}\right) / \partial I\right)$.
Iterative Gradient Sign Method (IGSM)~\cite{kurakin2017adversarial,madry2018towards} iteratively apply FGSM multiple times with a small step size to locate a stronger adversarial sample,
as shown in the following.
\begin{equation}
\begin{split}
& I_{t+1}^{'} = I_t - \alpha \cdot sign(\partial L(F(I_t,\theta),Y_{adv})/\partial I) \\
& I_{t+1} = clip(I_{t+1}^{'}, I - \epsilon, I + \epsilon)
\end{split}
\end{equation}
where $I_t$ is the adversarial example synthesized after $t$ iterations and $I_0 = I$. $I_{t+1}^{'}$ is a temporary variable. $clip(\cdot, l, u)$ with lower bound $l$ and upper bound $u$, is an element-wise operator that ensures the $L_\infty$ of the adversarial perturbation $|I_{t+1} - I|$ within the bound $\epsilon$.
Momentum based Iterative FGSM (MI-FGSM)~\cite{dong2018boosting} introduces a momentum term to stabilize the gradient descent and avoid the adversarial example stuck in poor local minima, which helps to synthesize more transferable adversarial examples.
Athalye et al.~\cite{athalye2018obfuscated} propose three tricks, Backward Pass Differentiable Approximation, Expectation over Transformation and Reparameterization, which have broken into most existing obfuscated gradients based defenses under a white-box attack setting.

\subsection{Gray-box Attack Setting}
Adversarial attacks have multiple settings, according to how much knowledge that adversaries can access. The settings consist of white-box, gray-box and black-box. Adversaries in white-box attacks can obtain all information about target models and defense methods. In black-box setting, adversaries do not know the architecture, the training data, the parameters of target classifiers and defense methods. Gray-box attacks have more than one definition.
Previous works have defined different gray-box adversarial attacks and developed defense methods under their own settings.
Taran et al.~\cite{taran2018bridging} define a gray-box setting, where the attacker knows target network architecture, training/testing data and the output label for each input, but has no knowledge of the network parameters and the defense mechanism. Zheng and Hong~\cite{zheng2018robust} utilize another gray-box setting, where attackers know the architecture of the target network and its defense strategy, but have no knowledge of their parameters. Different from the above settings where adversaries cannot obtain the network parameters, Guo et al.~\cite{guo2018countering} introduce the following gray-box setting: the adversary has access to the model architecture and the model parameters, but is unaware of defense strategies. In this paper we adopt the gray-box setting in \cite{guo2018countering}. It is a strong gray-box attack setting since both the architecture and parameters of target networks can be accessed by adversaries.

\subsection{Defense against Adversarial Attack}
Many methods that aim at protecting some neural model from adversarial examples are proposed recently~\cite{lu2017safetynet,guo2018countering,xie2018mitigating,song2018pixeldefend,samangouei2018defense-gan,liao2018defense,prakash2018deflecting,akhtar2018defense,metzen2017on,li2019rosa,he2019non,yang2019adversarial}.
SafetyNet~\cite{lu2017safetynet} incorporates a deep convolutional neural network with a RBF-SVM which converts the final ReLU outputs to discrete codes to detect adversarial examples. Guo et al.~\cite{guo2018countering} utilize bit-depth quantization, JPEG compression, total variance minimization and image quilting to destory or remove adversarial noise. Xie et al.~\cite{xie2018mitigating} resort to random resizing and random padding during inference stage to leverage the cue that many adversarial examples are not scale invariant. Pixel Defend~\cite{song2018pixeldefend} iteratively updates each pixel of an input image using a pretrained PixelCNN~\cite{denoord2016pixel} that is learned to predict a pixel value based on other pixels. Defend-GAN~\cite{samangouei2018defense-gan} learns to model the data distribution of clean images, and solves a vector in its learned latent space to approximate an input image. The latent vector is exploited to synthesize a substitute for the input via a generative neural network. HGD~\cite{liao2018defense} makes use of high-level feature extracted by the attacked target network to train an image denoiser that can remove adversarial noise. Since HGD requires accessing the intermediate outputs of target classifiers to tune the denoiser, it is not a portable defense.
MagNet~\cite{Meng2017MagNet} utilizes a detector to reject inputs far from the manifold boundary of training data, and an auto-encoder as reformer to find a substitute that is close to the input on the manifold. It is not straight-forward to fairly compare MagNet with other methods that do not reject adversarial examples. MagNet is based on the assumption that samples of some classification task are on a manifold of lower dimensions, and is only evaluated with small-size images. 
Pixel Deflection~\cite{prakash2018deflecting} iteratively and locally swaps two randomly sampled pixels according to their positions, before applying image denoiser to destroy adversarial noise. {\color{black}DIP~\cite{kattamis2019exploring} online trains a CNN that reconstructs the input image from a noise map. The output of the trained CNN is sent to be classified. Different from DIP, the proposed method updates an image and a CNN from scratch in an alternate way. The CNN does not directly output the synthesized image but approximates the energy of a data distribution for input images.
}
\section{Method}\label{sec:method}
This section describes the details of our proposed defense framework, online alternate generator, and its mathematical explanation. The proposed method can be regarded as a pre-processing component that protects a \textit{target network} or \textit{target classifier} from adversarial samples. \textit{Target network} is defined as a pre-trained neural model that is exposed to be attacked. The parameters of a target network can be accessed by attackers. That is to say, adversarial samples are synthesized with the target network. The proposed method is portable and practical, and it does not require accessing, modifying or re-training the parameters of the target network. Different from some existing pre-processing defenses that try to remove or destroy adversarial noises, our proposed method synthesizes an image from scratch.
The synthesized image is almost identical in appearance and semantics to the original image, but contains much less adversarial perturbations, and hence achieve more robust classification.
\begin{figure*}[!t]
	\begin{center}
		\includegraphics[width=1.0\linewidth]{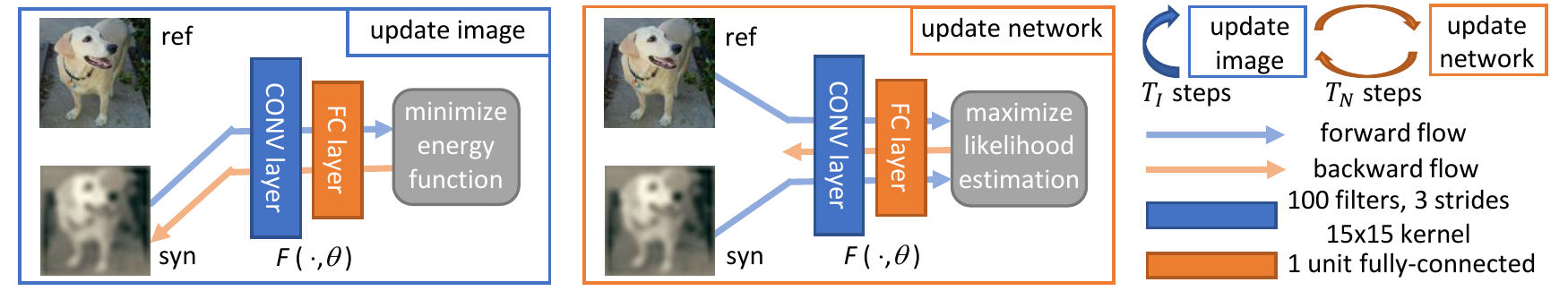}
	\end{center}
	\caption{A Defense Framework: Online Alternate Generator. ref denotes the reference image $I_z$. syn denotes the synthesized image $I_s$. $F(\cdot ,\theta)$ is the neural network with parameters $\theta$. The blue-boundary box represents image updating according to Eq. (\ref{Eq:updateImage}) while the orange-boundary box denotes network updating according to Eq. (\ref{updateTheta}). The blue bending arrow means the inner iterations of image updating in Algorithm~\ref{alg::OAG} while the orange bending arrows denotes the outer iterations of network updating.}
	\label{fig:framework}
\end{figure*}

The overall pipeline of the proposed method is described in Algorithm~\ref{alg::OAG}. Given an input image that may be an adversarial sample, we define \textit{reference image} (denoted as $I_z$ in Algorithm~\ref{alg::OAG}) as the input image, and synthesize another image $I_s$ to replace the original input before passing it into the target network for classification. For each input image, the parameters $\theta_1$ of the generator $F$ are initialized randomly and the synthesized image $I_1$ is filled with zeros at the beginning. $T_N$ denotes the maximum number of outer iterations while $T_I$ denotes the maximum number of inner iterations. Within each outer iteration, the synthesized image is updated for $T_I$ times while the network parameters of our proposed method is updated once.Training the generator and synthesizing the image are conducted alternately. For each input image, the generator is updated for exactly $T_N$ times while the synthesized image $I_s$ is updated for $T_N*T_I$ times.
Notice that $\theta_t$ in Algorithm~\ref{alg::OAG} does not include any parameters of the target network, but only the parameters of the proposed defense method.
In Line 7 of Algorithm~\ref{alg::OAG}, $I_{T_I+1}$ is assigned to $I_1$, since we employ a circular array $I_{\{1, ..., T_I+1\}}$ to restore $T_I+1$ latest snapshots of the synthesized image.
An overview of the proposed method is illustrated in Fig.~\ref{fig:framework}, it is composed of an image updating procedure and a network updating process. Please refer to~Algorithm~\ref{alg::OAG} for the detailed  iteration. In the following sections, we will focus on the  computation of both image updating and network updating, and the theoretical basis behind this alternative updating mechanism.


\renewcommand{\algorithmicrequire}{\textbf{Input:}}
\renewcommand{\algorithmicensure}{\textbf{Output:}}
\begin{algorithm}[h]
	\caption{Online Alternate Generation Algorithm}
	\label{alg::OAG}
	\begin{algorithmic}[1]
		
		\Require
		$I_z$, reference image with potential adversarial noise
		\Ensure
		$I_s$, synthesized image
		\State Randomly initialize $\theta_1$
		\State $I_1 = 0$
		
		\For{$t=1$ to $T_N$}
		\For{$s=1$ to $T_I$}
		\State Update $I_{s+1}$ with $I_s$ according to Eq.~(\ref{Eq:updateImage})
		\EndFor
		\State $I_1 = I_{T_I+1}$
		\State Update $\theta_{t+1}$ with $\theta_t$ according to Eq.~(\ref{updateTheta})
		\EndFor
		\State {$I_s = I_{T_I+1}$}
		\State \textbf{return} $I_s$
	\end{algorithmic}
\end{algorithm}

\subsection{Image Updating}\label{sec:imageupdating}
Let us discuss how to update a synthesized image initialized by zeros, with a reference image $I_z$ and a neural network $F$.
$F$ denotes the proposed generator instead of the target classifier.
Suppose $I_z$ and $I_s$ are sampled from the same data distribution denoted as $p(I;\theta) = (1/Z) e^{-U\left(I;\theta\right)}$ where $I$ denotes an image. $\theta$ represents the parameters of the model and $Z$ is a normalization term. $e$ denotes exponential and $U$ is an energy function.
Then we utilize the neural network $F$ in the proposed framework to approximate the energy function, i.e., $F(I,\theta) = -U\left(I;\theta\right)$. To maximize the probability density of the synthesized image $I_s$, we update $I_s$ to minimize the energy function by $I_{s+1} = I_s - \alpha \partial U\left(I_s;\theta\right)/ \partial I = I_s - \alpha \left(- \partial F\left(I_s,\theta\right)/\partial I \right)$ where $I_s$ is the current synthesized image and $I_{s+1}$ is the updated image. $\alpha$ denotes the learning rate.
$\partial F(I,\theta) / \partial I$ is the gradients of neural network $F(\cdot,\theta)$ with respect to image $I$, and can be computed by backward propagation. In a sense, generating an image is to reconstruct the reference image. However, synthesizing an image with adversarial noise is undesirable. Thus we further introduce a noise model during image updating,
\begin{equation} \label{random_sgd}
I_{s+1} = I_s - \alpha \left(- \partial F\left(I_s,\theta\right) / \partial I \right) + \epsilon_g D
\end{equation}
{\color{black}where $D$ denotes some noise distribution, such as a gaussian noise $N(0,1)$. $\epsilon_g$ is the strength of the noise.} Adding noise during image synthesis can increase the difficulty in recovering subtle details, and thus decrease the chance of fitting adversarial noise. {\color{black}Langevin Dynamics, which is originally to simulate how particles move under a random force, has a similar form with Eq~(\ref{random_sgd}), a modified gradient decent with a Gaussian perturbation. To comprehend the relationship between $\alpha$ and $\epsilon_g$, we resort to Langevin Dynamics and come up with Eq~(\ref{Eq:updateImage}) following a previous work~\cite{lu2016learning}.}
\begin{equation}\label{Eq:updateImage}
I_{s+1} = I_s - (\epsilon_g^2 / 2) \left(I_s - \partial F\left(I_s,\theta\right) / \partial I \right) + \epsilon_g N\left(0,1\right)
\end{equation}
{\color{black}where $\epsilon_g$ controls the magnitude of the Gaussian noise.} $\epsilon_g^2 / 2$ corresponds to the learning rate $\alpha$. $1 - \epsilon_g^2 / 2$ is the inertia factor of $I_s$. Since random fluctuation is used to generate an image, the distribution of images is changed into
$p(I;\theta) = (1/Z) e^{-U(I;\theta)} (1/(2\pi)^{|S|/2}) e^{- \frac{1}{2}||I||^2}$.
The multiplicative term on the right is a Gaussian distribution with $\sigma^2 = 1$. $|S|$ denotes the number of elements in image $I$.

\subsection{Network Updating}\label{sec:networkupdating}
The following details how to update the neural network $F$ such that the synthesized image $I_s$ gradually approximates the reference image $I_z$. Notice that at the very beginning $F$ is initialized by randomization. Thus we update $F$ to maximize the likelihood with respect to $I_z$. Let $L(\theta) = log( p(I_z;\theta) )$.
$\theta$ is trained along the direction maximizing the log likelihood $L(\theta)$ with gradient descent: $\theta_{t+1} = \theta_t + \beta \partial L(\theta_t) /\partial \theta$, where $\theta_t$ denotes the current parameters at the time step $t$. $\theta_{t+1}$ is the updated parameters. $\partial L(\theta_t) / \partial \theta$ denotes the gradients of the log likelihood function w.r.t $\theta$. 
{\color{black}As suggested in~\cite{xie2015learning}, the gradient is computed in:
\begin{equation}
\partial L(\theta) / \partial \theta = \partial F(I_z;\theta_t) / \partial \theta - E_{p(I;\theta)}[ \partial F(I;\theta_t) / \partial \theta ],
\end{equation}
where $E_{p(I;\theta)}[\cdot]$ is the expectation with $I$ following the distribution $p(I;\theta)$.} The expectation is not explicitly calculated but approximated by sampling. Langevin Dynamics, which is adopted to update the image, is also a tool to sample image $I$ from distribution $p(I;\theta)$. Thus we choose $\partial F(I_s;\theta_t) / \partial \theta$ to approximate the expectation $E_{p(I;\theta)}[\cdot]$ for simplicity. Then learning neural network $F$ can be formulated as in Equation~\ref{updateTheta}.
\begin{equation} \label{updateTheta}
\theta_{t+1} = \theta_t + \beta (\partial F(I_z;\theta) / \partial \theta - \partial F(I_s;\theta) / \partial \theta )
\end{equation}
{\color{black}For a testing image, $T_N$ samples $\partial F(I_s;\theta)/\partial \theta$ are selected to approximate $E_{p(I;\theta)}[\partial F(I;\theta)/\partial \theta]$. In practice, $T_N$ ranges from 200 to 300. Thus, hundreds of samples $\partial F(I_s;\theta)/\partial \theta$ are used to approximate the expectation $E_{p(I;\theta)}[\partial F(I;\theta)/\partial \theta]$ and such approximation is effective.}

\subsection{Analysis}\label{sec:analysis}
This subsection presents explanation on why the synthesized image $I_s$ approximate the reference image $I_z$ after alternately updating $I_s$ and the network $F$.
{\color{black}In the following, we consider $F$ as a very simple prototype. The prototype model contains a convolution layer, a ReLU layer and a summation operator. The prototype model contains a convolution layer, a ReLU layer and a summation operator. $I$, the input of prototype model, could be reshaped as a vector of shape $1\times c h w$ . The convolution layer has $K$ kernels and the kernel size is $r_h\times r_w$. The weight of the convolution is denoted as $W$ of size $cr_hr_w\times K$, while the bias $B$ is of size $1\times K$. For simplicity, we let $h=r_h$ and $w=r_w$ so that the convolution operator is only applied on a position. The output of the convolution is $IW+B$, of size $1\times K$. The ReLU function $\lambda$ is applied on each element of $IW+B$. The summation operator sums up all elements of $\lambda(IW+B)$. Thus $F(I)$ outputs a single scalar and is formulated as:}
\begin{equation} \label{FI}
F(I) = \sum\nolimits_{k=1}^K \lambda(IW_k+B_k)
\end{equation}

According to the definition of ReLU, $\lambda(x) = max(0, x)$. ReLU function can be represented as a multiplication between the input and a Heaviside step function $u$. For $x > 0$, $u(x)$ equals 1. Otherwise, $u(x)$ is 0. Thus $\lambda(x) = u(x)x$. Then we formulate Eq~(\ref{FI}) as:
\begin{equation}
\begin{split}
F(I) &= \sum\nolimits_{k=1}^K (u(IW_k+B_k)(IW_k+B_k)) \\
&= \sum\nolimits_{k=1}^K (u(I,\theta)(IW_k+B_k)),
\end{split}
\end{equation}
where $u(IW_k+B_k)$ is a single scalar depending on $I$ and $\theta$. We denote $u(IW_k+B_k)$ as $u(I,\theta)$ for simplicity.

{\color{black}Assume that the synthesized image $I_s$ and the reference image $I_z$ belong to the same data distribution $p(I;\theta)$. Since we introduce Gaussian noise to synthesize $I_s$, we assume that 
\begin{equation}
p(I;\theta) = \frac{1}{Z}e^{-U(I;\theta)} \frac{1}{(2\pi)^{|S|/2}}e^{-||I||^2/2}
\end{equation}
Let $I^*$ denote the image that maximizes the probability $p(I;\theta)$. The proposed method is to synthesize an image $I_s$ that approximates to $I^*$. Let $F(I;\theta)$ approximate $-U(I;\theta)$. To maximize the probability, we need to maximize $-U(I;\theta)-||I||^2/2$ as:}
\begin{equation} \label{prob}
\begin{split}
&F(I)-||I||^2/2 \\
= &-||I||^2/2 + I\sum\nolimits_{k=1}^K(u(I,\theta)W_k) + \sum\nolimits_{k=1}^Ku(I,\theta)B_k \\
= &-||I - \sum\nolimits_{k=1}^K(u(I,\theta)W_k)||^2/2 + C(u(I,\theta),\theta)
\end{split}
\end{equation}
where $C(u(I,\theta),\theta)$ depends on $I$ and $\theta$. When we update image $I_s$ as $I_{s+1}$, we can set $u(I,\theta)$ as $u(I_s,\theta)$ and fix the network parameters $\theta$. {\color{black}Then we come up with $I^* = \sum\nolimits_{k=1}^K(u(I_s,\theta)W_k)$ to maximize Eq~(\ref{prob}). Recall how we update $I_{s+1}$ as shown in Eq~(\ref{Eq:updateImage}):
\begin{equation} \label{linear_interp}
\begin{split}
I_{s+1} &= I_s - (\epsilon_g^2/2)(I_s - \partial F(I_s,\theta)/\partial I) + \epsilon_gN(0,1)\\
&= (1-\epsilon_g^2/2)I_s + \epsilon_g^2/2\cdot \partial F(I_s,\theta)/\partial I + \epsilon_gN(0,1)\\&= (1-\epsilon_g^2/2)I_s + \epsilon_g^2/2 \sum\nolimits_{k=1}^K(u(I_s,\theta)W_k) + \epsilon_gN(0,1)\\
&= (1-\epsilon_g^2/2)I_s + \epsilon_g^2/2\cdot I^* + \epsilon_gN(0,1)
\end{split}
\end{equation}
As Eq $(v)$, updating $I_{s+1}$ is to do linear interpolation between $I_s$ and $I^*$ with a Gaussian perturbation. Thus the pixel values of the synthesized image $I_s$ will approximate to those of $I^*$, which is the peak of highest probability. Since $I_z$ is a sample of high probability in the same data distribution with $I^*$, $I_s$ will also be visually similar to $I_z$. When updating the network, the proposed method tunes $\theta$ to guarantee that the probability of sampling $I_s$ is as high as sampling $I_z$.}


In our real implementation (shown in Fig.~\ref{fig:framework}) of the proposed method, the neural network $F$, which contains a convolution, a fully-connected layer and a non-linear activation, can be approximated by the above-mentioned model. Therefore the above analysis also works for our actual implementation. 
{\color{black}
Here we discuss the time complexity of the proposed online alternate generation algorithm (shown in Algorithm~\ref{alg::OAG}). The proposed algorithm consists of two nested loops. The outer loop contains $T_N$ iterations while the inner loop has $T_I$ iterations. Assume that computing Eq.~(\ref{Eq:updateImage}) takes the same constant time as computing Eq.~(\ref{updateTheta}). The overall time complexity is $\mathcal{O}(T_N(T_I+1)) = \mathcal{O}(T_N\cdot T_I)$. In practice, it takes about 30 seconds to process an input image of size $224\times 224$.
}

\subsection{Stopping Criteria}
The proposed online alternate generation terminates after $T_N$ outer iterations as shown in Algorithm~\ref{alg::OAG}. On one hand, larger network step $T_N$ leads to unnecessary computations. On the other hand, with smaller $T_N$, the proposed generator could fail to reproduce the semantics of an input image, which will seriously drop the classification accuracy of target networks. More importantly, to develop a portable defense against unseen attacks, we need to determine $T_N$ without tuning it on known adversarial samples. Here, we propose to utilize images perturbed by Gaussian noises to choose the network steps $T_N$. It is based on an assumption that the online alternate generating process of adversarial samples and those with Gaussian noises are similar. The experimental details and results are presented in Section~\ref{sec:abla_study}.

\section{Experiments}
\subsection{Experimental Setting}\label{sec:exp_setting}
\begin{table*}[!t]
	\centering
	\tiny
	\caption{Top-1 accuracy of different defense methods against FGSM, IGSM, MI-FGSM and C\&W on ILSVRC 2012 dataset and Oxford Flower-17 dataset. 
  $*$ indicates that the method needs to be trained before inference.}
	\resizebox{\linewidth}{!}{
		\begin{tabular}{l|rrrr|rrrr}
			\toprule
			& \multicolumn{4}{c|}{\textbf{ILSVRC 2012}} & \multicolumn{4}{c}{\textbf{Oxford Flower-17}} \\
			\multirow{-2}{*}{\textbf{Defense Methods}} & FGSM & IGSM & MI-FGSM & C\&W & FGSM & IGSM & MI-FGSM & C\&W \\
			\midrule
			None & 8.35\% & 0.05\% & 0.30\% & 0.00\% & 27.65\% & 7.35\% & 0.59\% & 0.00\% \\
			Mean Filter \cite{li2017adversarial} & 39.65\% &  {\mybluecolor 67.65\%} & 38.50\% & {\mybluecolor 75.55\%} & 60.59\% & 73.24\% & 47.06\% & 66.47\% \\
			JPEG \cite{Dziugaite2016A} & 19.70\% & 54.55\% & 0.85\% & 67.50\% & 45.88\% & 47.94\% & 20.88\% & 28.82\% \\
			TVM \cite{guo2018countering} & {\mygreencolor 41.00\%} & 66.70\% & {\mybluecolor 53.45\%} & {\mygreencolor 69.00\%} & 69.12\% & {\mybluecolor 83.82\%} & {\mybluecolor 70.88\%} & {\mybluecolor 80.29\%} \\
			Pixel Deflection \cite{prakash2018deflecting} & {\mybluecolor 41.70\%} & 55.80\% & {\mygreencolor 51.10\%} & 57.00\% & {\mygreencolor 71.47\%} & 82.06\% & {\mygreencolor 70.59\%} & {\mygreencolor 78.82\%} \\
			Randomization \cite{xie2018mitigating} & 40.05\% & {\mygreencolor 67.15\%} & 44.80\% & 68.30\% & {\mybluecolor 72.65\%} & {\mygreencolor 83.24\%} & 62.94\% & 75.88\% \\
			$*$Pixel Defend \cite{song2018pixeldefend} & 20.15\% & 55.25\% & 20.05\% & 66.75\% & 48.82\% & 53.53\% & 27.06\% & 37.35\% \\
			Ours & {\color{red} 49.10\%} & {\color{red} 75.25\%} & {\color{red} 55.90\%} & {\color{red} 77.95\%} & {\color{red} 76.76\%} & {\color{red} 87.35\%} & {\color{red} 73.82\%} & {\color{red} 84.41\%} \\
			\bottomrule
		\end{tabular}
	}
	\label{table:comparision}
\end{table*}
\paragraph{Dataset}
We evaluate the performance of our method on ILSVRC 2012 dataset \cite{Deng2009ImageNet} and Oxford Flower-17 dataset \cite{Nilsback06}. Most images in ILSVRC 2012 dataset are large images whose size is not smaller than $200 \times 200$. Most images in Oxford Flower-17 dataset are larger than $500 \times 500$. We claim that it is practical and meaningful to develop a defense method that works with large images. Because experimental results obtained on simple dataset such as MNIST~\cite{lecun1998gradient} do not always generalize to harder tasks, as suggested by \cite{tramer2018ensemble}. 
As suggested in \cite{xie2018mitigating}, it is less meaningful to attack misclassified images, we randomly choose 2000 correctly classified images (2 images/class) from the validation set to perform experiments.

\paragraph{Target Network}
We choose ResNet18 \cite{He2015Deep} as the attacked target network on Oxford FLower-17 dataset, and utilize ResNet50 on ILSVRC 2012 dataset. To demonstrate the transferability of out proposed defense method, we further conduct experiments on Oxford Flower-17 dataset with different target networks, including VGG11 \cite{simonyan2015very}, MobileNet\_v2 \cite{sandler2018mobilenetv2} and DenseNet121 \cite{huang2017densely}.

\begin{figure}[!t]
	\centering
	\begin{center}
		\includegraphics[width=1.0\linewidth]{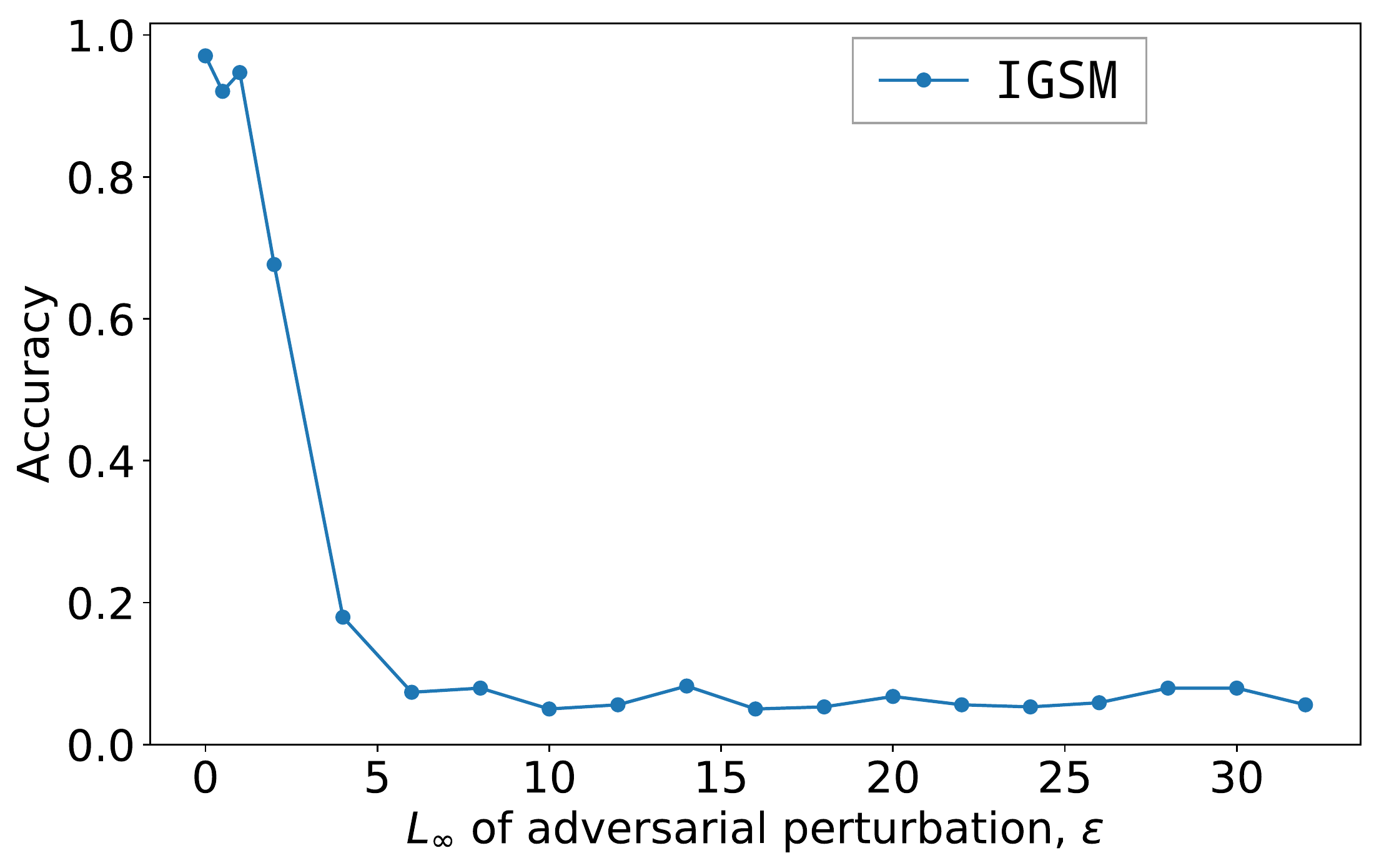} 
	\end{center}
	\caption{Top-1 accuracy of ResNet18 attacked by targeted IGSM with different $\epsilon$ ($l_\infty$ norm of adversarial perturbations) on Oxford Flower-17 dataset.}
	\label{chart:attack_strength_selection}
\end{figure}

\paragraph{Attack Methods}
We exploit FGSM \cite{goodfellow2015explaining} and IGSM \cite{kurakin2017adversarial} to construct targeted adversarial examples with randomly selected target categories. We utilize MI-FGSM~\cite{dong2018boosting} and C\&W attack~\cite{carlini2017towards} to synthesize untargeted adversarial samples.
In this paper, we adopt gray-box attacks~\cite{guo2018countering} in which attackers can access the target network and its parameters, but have no knowledge of defense methods. $L_\infty$ norm is adopted to bound adversarial perturbations. $\epsilon$ denotes the upper bound of $L_\infty$ norm. We choose $\epsilon$ for each attack type such that the adversarial attack is strong enough and its resulting perturbation is imperceptible.
For example, we attack a ResNet18 model by targeted IGSM, as shown in Fig.~\ref{chart:attack_strength_selection}. As $\epsilon$ grows from 0 to 5, the top-1 accuracy of the attacked model drops rapidly. For $\epsilon$ larger than 5, the top-1 accuracy becomes stable. In such case, larger perturbations do not lead to stronger attack but only synthesize noisy and undesirable images. 
For FGSM, IGSM and MI-FGSM, $\epsilon$ is selected respectively as \{6, 6, 2\} on ILSVRC 2012 dataset and \{6, 6, 4\} on Oxford Flower-17 dataset. The number of iterations is set as min($\epsilon$+4, ceil(1.25$\epsilon$)), according to~\cite{kurakin2017adversarial}. ceil($\cdot$) denotes rounding up to an integer. The step size is set as 1.

\paragraph{Defense Methods}
In this paper, we compare our method with six state-of-the-art portable defenses, including Mean Filter \cite{li2017adversarial}, JPEG compression and decompression \cite{Dziugaite2016A}, TVM \cite{guo2018countering}, Pixel Deflection \cite{prakash2018deflecting}, Randomization \cite{xie2018mitigating} and Pixel Defend \cite{song2018pixeldefend}. All of these methods play a role as pre-processings. In addition, the first five methods can be used without training. Although Pixel Defend needs to train a PixelCNN \cite{denoord2016pixel} network on the training set, it does not need to access the parameters of adversarial attacks. Pixel Defend has been successfully attacked by Athalye et al.~\cite{athalye2018obfuscated} in white-box setting, but it is still meaningful to compare it with our proposed method in the gray-box setting.

\subsection{Comparison with the State-of-the-art}
We choose existing portable defenses that can work with unseen attack types and large images for comparison. These defense methods do not require accessing target networks or attack types. This experiment is conducted with gray-box attack setting. The parameters of all defenses are invisible to attackers. As can be observed in TABLE \ref{table:comparision}, our proposed defense method achieves the best top-1 accuracy on both ILSVRC 2012 dataset and Oxford Flower-17 dataset, against four kinds of attacks, including FGSM, IGSM, MI-FGSM and C\&W. On ILSVRC 2012 dataset, The proposed method outperforms the second best Pixel Deflection by 7.4\% against FGSM, and surpasses the second best Mean Filter by 7.6\% against IGSM.
Our proposed method outperforms the second best method TVM by 2.45\% top-1 accuracy on MI-FGSM and Mean Filter by 2.4\% top-1 accuracy on C\&W.
On Oxford Flower-17 dataset, the top-1 accuracy of our method is 4.1\% higher than the second best against FGSM and 3.5\% higher than TVM against IGSM. The proposed method outperforms the second best method, TVM, by 2.94\% top-1 accuracy on MI-FGSM and 4.12\% top-1 accuracy on C\&W attack.

TABLE~\ref{table:comparision} shows that top-1 accuracy on IGSM is better than FGSM, while the results in \cite{madry2018towards} suggest that IGSM is stronger than FGSM on white-box attacks with the same $\epsilon$. We claim that our experimental results is reasonable because our experiments is tested on gray-box attacks. In our cases, defense methods shown in TABLE~\ref{table:comparision} have modified the adversarial perturbations in adversarial samples. Since adversarial perturbations computed in iterative methods `fit' the target network better than single-step methods, minor changes on iterative adversarial samples could reduce more attack effects than single-step adversarial samples. It can be regarded as an evidence supporting that stronger adversaries decrease transferability~\cite{madry2018towards}. Experimental results in Pixel Deflection~\cite{prakash2018deflecting} also show that classifier accuracy of IGSM is higher than FGSM on gray-box attacks. Notice that the numeric results of Pixel Deflection in our experiment may differ from those in~\cite{prakash2018deflecting}. It is due to that we adopt $L_\infty$ norm following the original FGSM/IGSM while Prakash et al.~\cite{prakash2018deflecting} use $L_2$ norm. Different norm may result in different distribution of adversarial noises and attack effects.

{\color{black}We discuss two reasons why the proposed method are better than the previous works. First, the proposed method synthesizes a new image with less adversarial noises to replace the original input. In Sec III-C we have shown that the synthesized image $I_s$ approximates to the original input $I_z$. The introduced Gaussian perturbation in Eq~(\ref{Eq:updateImage}) avoids recovering the adversarial noises of $I_z$. Second, the transferability among CNN models may be a reason. When updating image $I_s$, $F$ has also been updated for hundreds of times. That is to say, $I_s$ is synthesized based on hundreds of different CNN models (a model $F$ with different parameters). The pixel values of $I_s$ could be suitable for another CNN model (such as the target classifier) to extract effective features.}

\subsection{Investigation with Adversarial Training}
\begin{table}[!t]
	\centering
	\caption{Investigation of the proposed method with Ensemble Adversarial Training. Adversarial examples are synthesized using MI-FGSM on ILSVRC 2012 dataset.}
	\label{table:with_ensemble}
	\resizebox{\linewidth}{!}{
		\begin{tabular}{l rrr}
			\toprule
			\textbf{Attacks With} & IncV3ens & IncV3 + ours & IncV3ens + ours \\
			\midrule
			IncV4 & 71.80\% & 73.30\% & 78.45\% \\
			IncV3 & 34.70\% & 39.60\% & 71.70\% \\
			\bottomrule
		\end{tabular}
	}
\end{table}
This section presents how our proposed method work with Ensemble Adversarial Training method~\cite{tramer2018ensemble}. Noted that adversarial training based methods are not \textit{portable} defense as they require to re-train the attacked target networks. As shown in TABLE~\ref{table:with_ensemble}, IncV3ens, IncV3+ours and IncV3ens+ours respectively denote an ensemble adversarial training IncV3 (Inception-V3)~\cite{szegedy2016rethinking} model, a plain IncV3 model defended by our proposed method and the Ensemble Adversarial Training model defended by our proposed method. `Attacks With' denotes the pre-trained models used to synthesize adversarial examples. The pre-trained IncV3 model has the same architecture but different parameters from IncV3ens. Thus attacks with IncV4 (Inception-V4)~\cite{szegedy2017inception} and IncV3 are in black-box setting and gray-box setting respectively. The results indicate that the proposed method outperforms Ensemble Adversarial Training by 1.5\% and 4.9\% top-1 accuracy in black-box and gray-box setting respectively. Besides, our proposed method significantly enhances the top-1 accuracy of the IncV3ens model by 6.65\% and 37\% top-1 accuracy against attacks with IncV4 and IncV3 respectively. {\color{black}Notice that Ensemble Adversarial Training and the proposed method obtain a relatively lower accuracy (less than 40\%) on gray-box setting. Such situation could be due to that the IncV3 model is relatively more sensitive to the MI-FGSM attack than other models such as ResNet. Besides, we find that combining the proposed method with Ensemble Adversarial Training can achieve acceptable top1-accuracy of 71.70\%.}

\subsection{Transferability Analysis}\label{sec:trans_analysis}
\begin{table}[!t]
	\centering
	\Large
	\caption{Top-1 accuracy of different defense methods against IGSM on Oxford Flower-17 dataset with different target networks. 
  $*$ indicates that the method requires offline training.}
	\resizebox{\linewidth}{!}{
	\begin{tabular}{l rrrr}
		\toprule
		\textbf{Defense Methods} & \textbf{ResNet18} & \textbf{VGG11}  & \textbf{MobileNet\_v2} & \textbf{DenseNet121} \\
		\midrule
		None & 7.35\% & 9.71\% & 5.59\% & 5.29\% \\
		Mean Filter & 73.24\% & 69.12\% & 82.65\% & 60.00\% \\
		JPEG  & 47.94\% & 50.00\% & 58.53\% & 27.56\% \\
		TVM   & {\mybluecolor 83.82\%} & 67.94\% & {\mybluecolor 88.53\%} & {\mygreencolor 73.24\%} \\
		Pixel Deflection  & 82.06\%& {\mybluecolor 77.06\%} & 82.65\% & {\mybluecolor 82.56\%} \\
		Randomization  & {\mygreencolor 83.24\%} & {\mygreencolor 72.65\%} & {\mygreencolor 86.75\%} & 72.53\% \\
		$*$Pixeldefend  & 53.53\% & 50.88\% & 61.47\% & 35.29\% \\
		Ours & {\color{red} 87.35\%} & {\color{red} 80.29\%} & {\color{red} 91.18\%} & {\color{red} 87.65\%}  \\
		\bottomrule
	\end{tabular}%
	}
	\label{table:transferability}
\end{table}

This section investigates whether our defense method is transferable to protect different kinds of target networks. Experiments are conducted on Oxford Flower-17 dataset. Adversarial examples are generated using IGSM in this section. The target networks includes VGG11, MobileNet\_v2 and DenseNet121. These target networks are initialized with weights pretrained on ImageNet, and then fine-tuned on the training set of Oxford FLower-17 dataset. The top-1 accuracy of VGG11, MobileNet\_v2 and DenseNet121 are 91.47\%, 97.35\% and 96.47\% respectively on clean images from the test set. We determine $\epsilon$ according to the criteria discussed in the above section, and select $\epsilon=6$ for ResNet18, $\epsilon=8$ for VGG11, $\epsilon=6$ for MobileNet\_v2 and $\epsilon=8$ for DenseNet121.

As shown in TABLE~\ref{table:transferability}, the proposed method demonstrates higher top-1 accuracy than other state-of-the-art defense algorithms. Our method exceeds TVM by 3.5\% top-1 accuracy on ResNet18, and outperforms Pixel Deflection by 3.2\% top1-accuracy on VGG11. On MobileNet\_v2, the proposed method surpasses the second best TVM by 2.6\% top-1 accuracy. The top1-accuracy of our method is 5.1\% higher than the second best Pixel Deflection on DenseNet121. Notice that the proposed method does not need offline training beforehand, and therefore does not obtain any knowledge or response of the attacked target networks. Tuning hyper-parameters is also independent of attacked target models. Thus the proposed method is not biased towards any specific models, but enjoys superior transferability with various types of target networks.

\subsection{Ablation Study}\label{sec:abla_study}
\begin{figure}[t]
	\centering
	\begin{center}
		\includegraphics[width=1.0\linewidth]{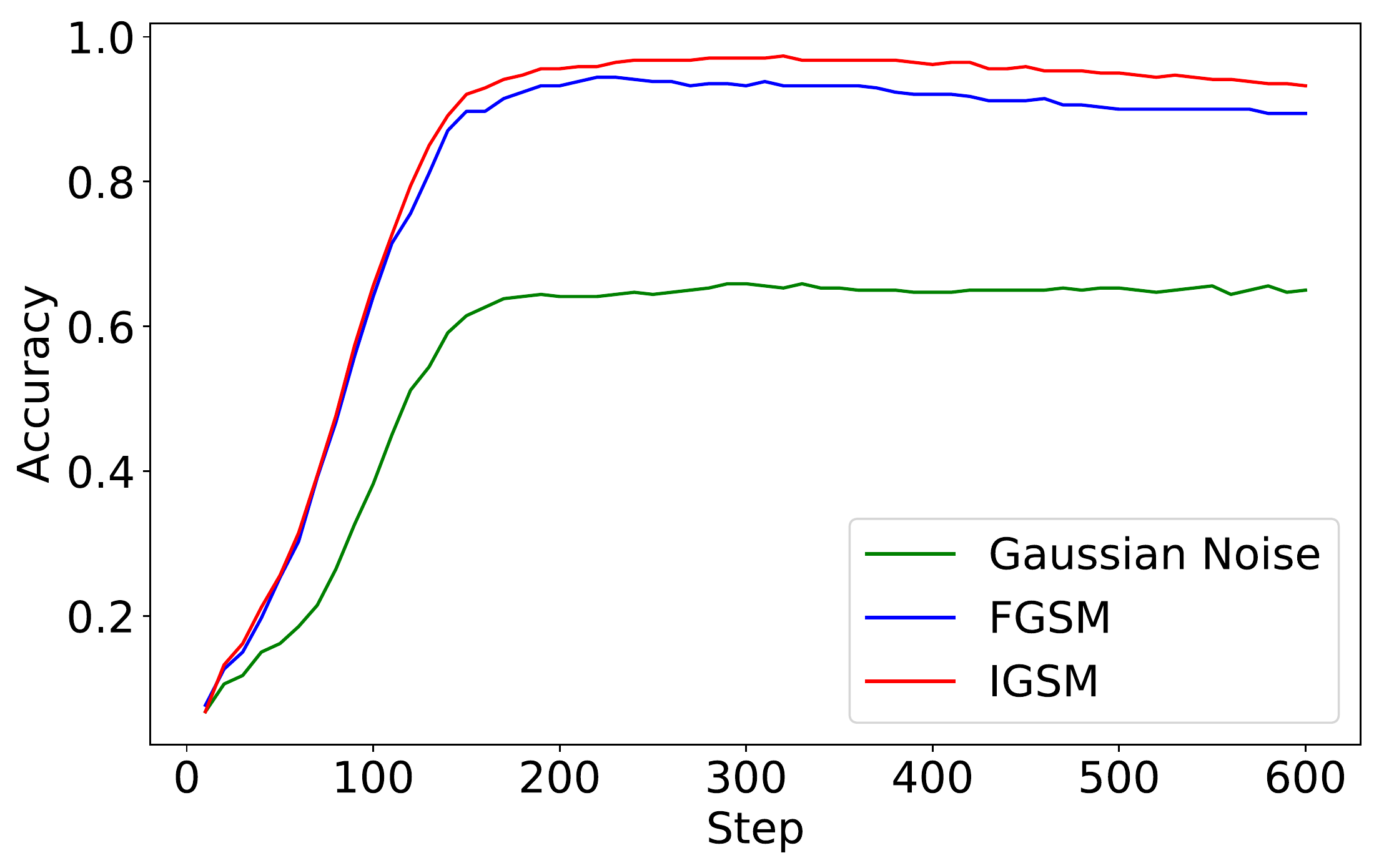}	
	\end{center}
	\caption{Comparison among reference images with different noises. `Accuracy' denotes top-1 accuracy. `Step' corresponds to Network Steps. \textbf{Gaussian Noise} means that images are degraded by Gaussian Noise.  \textbf{FGSM} and \textbf{IGSM} denote adversarial noises generated by FGSM and IGSM. }
	\label{fig:Ablation}
\end{figure}

Our proposed defense method contains several essential hyper-parameters, including Network Steps $T_N$, Image Steps $T_I$ and kernel size. We investigate how different settings of these hyper-parameters affect the performance of our method on Oxford Flower-17 dataset. When investigating a hyper-parameter, other hyper-parameters are set as default values suggested in Section~\ref{sec:exp_setting}.

\paragraph{Network Steps $T_N$}
{\color{black}To determine $T_N$, the number of iteration (in the outer loop), a straightforward way is to generate adversarial samples with training set, run the proposed method on these adversarial samples, and find out a suitable value for $T_N$. However, as a defender we usually do not know the exact attack type in testing stage. Thus it is meaningful to determine $T_N$ without the knowledge of adversarial attacks. Since our proposed method does not estimate the attack type of an input image, we assume that using samples with non-adversarial noises also can determine the iteration number. Then we conduct an experiment to verify the assumption. We generate three sets of noisy samples for training set. One set is generated by adding Gaussian noises ($\mu=0$, $\sigma=\pm0.25$). The other two sets are generated by adding adversarial noises of FGSM and IGSM. We run the proposed method for different iterations on these three sets of data. We use the target image classifier to classify the images synthesized by the proposed method at different iterations. The results is shown in Fig~\ref{fig:Ablation} that has taken average of the whole training set. It does reflect most of cases. 
The resulting curve of IGSM, FGSM and Gaussian noises are marked with red, blue and green color in Fig~\ref{fig:Ablation}. 
As can be seen, these three curves have almost the same trend, which converge and become flat after around 200 steps. Thus we can barely rely on the curve corresponding to Gaussian noise to determine a suitable Network Steps. On ILSVRC 2012 dataset, Network Steps is selected as 600 in the same way.}

\paragraph{Image Steps $T_I$}
\begin{table}[!t]
	\centering
	\caption{Comparison among different Image Steps.\iffalse Adversarial examples are synthesized on Oxford Flower-17 dataset using IGSM.\fi}
	\resizebox{\linewidth}{!}{
		\begin{tabular}{l rrrr}
			\toprule
			\textbf{Image Steps $T_I$} & 10 & 20 & 30 & 40 \\
			\midrule
			\textbf{Top-1 Accuracy} & 87.05\% & 87.35\% & 87.35\% & \textbf{87.65\%} \\
			\bottomrule
		\end{tabular}
	}
	\label{table:ablation_on_flower}
\end{table}
\begin{figure*}[!t]
	\begin{minipage}[t]{0.5\linewidth}
		\centering
		\includegraphics[width=1\linewidth]{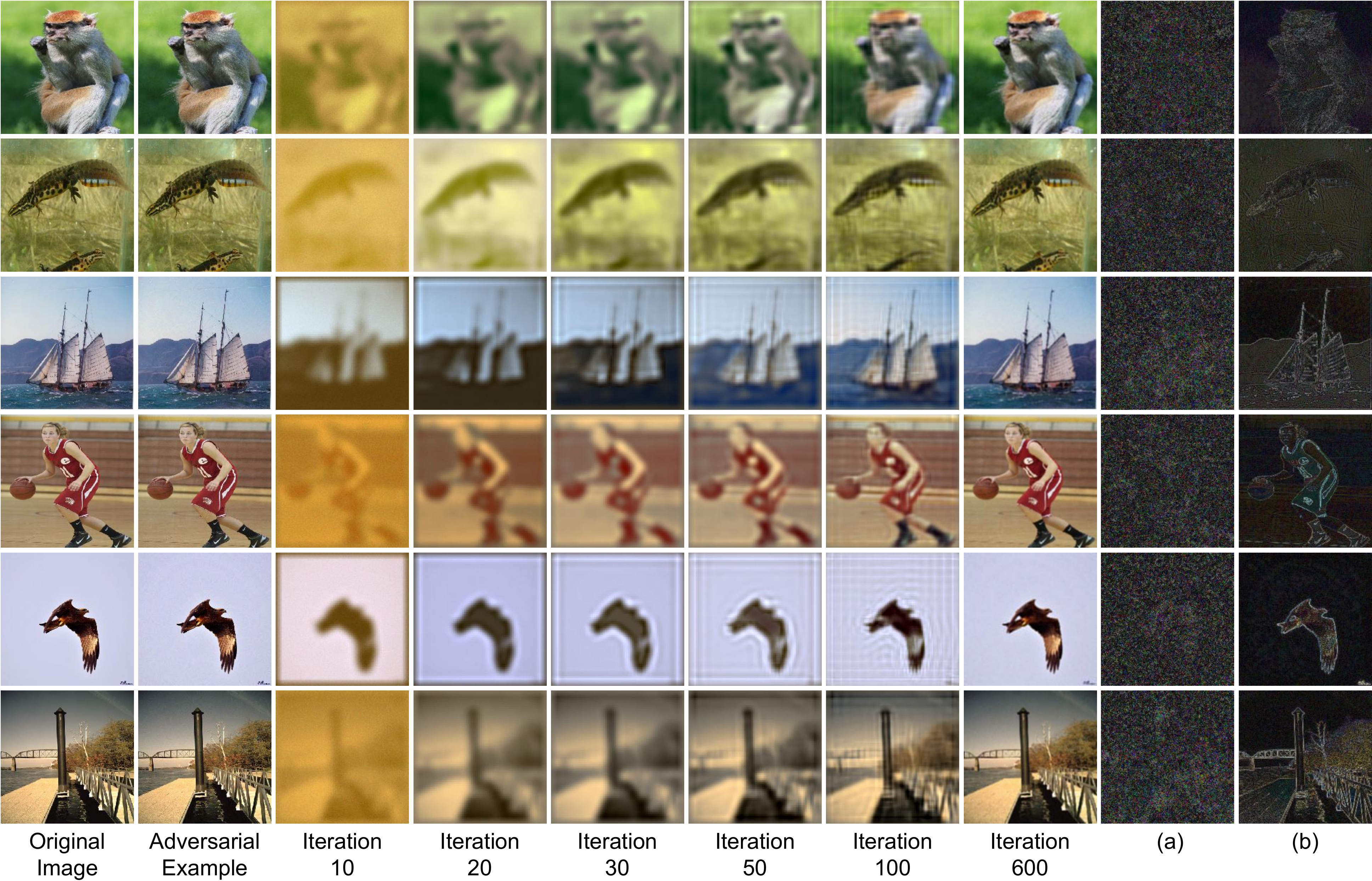}
		i). ILSVRC 2012 dataset
	\end{minipage}
	\begin{minipage}[t]{0.5\linewidth}
		\centering
		\includegraphics[width=1\linewidth]{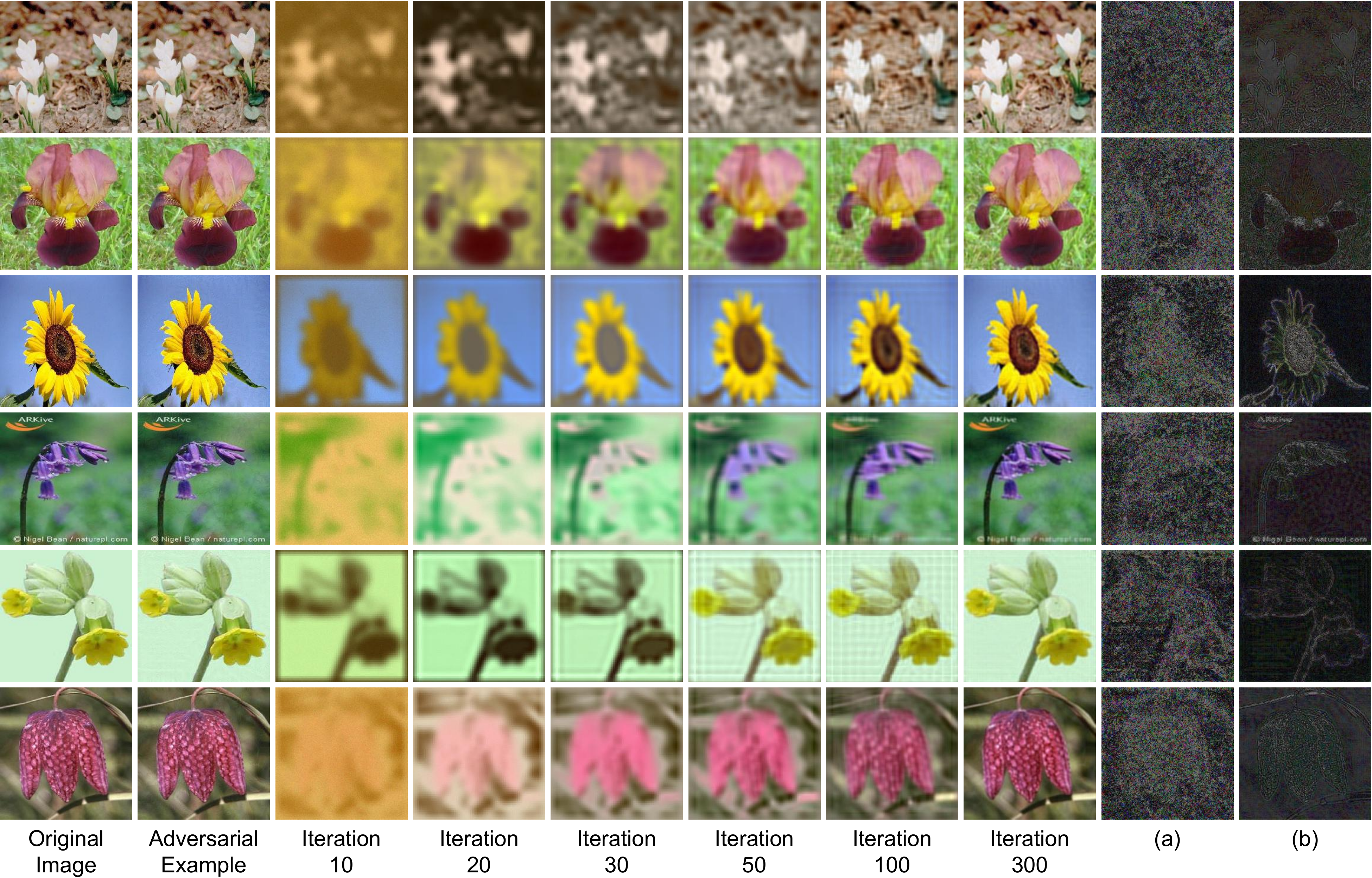}
		ii). Oxford Flower-17 dataset
	\end{minipage}
	\caption{The visualization of our defense method on ILSVRC 2012 and Oxford Flower-17 dataset under targeted Iterative FGSM attack. We exhibit images generated by our defense method at different iterations. (a) visualizes the difference between adversarial examples and original images. (b) visualizes the difference between our results and original images. As can be observed, our results have different noise distribution from input adversarial examples.}
	\label{fig:ILSVRC_Flower}
\end{figure*}
We evaluate four different Image Steps with kernel size set to $15\times15$ and Network Steps set to 300. As shown in TABLE~\ref{table:ablation_on_flower}, the top-1 accuracy increases as more Image Steps is adopted. However, as larger Image Steps leads to higher time cost, we select $T=20$ to make a trade-off between performance and efficiency.

\paragraph{Kernel Size}
\begin{table}[!t]
	\centering
	\caption{Comparison among different kernel sizes.\iffalse Adversarial examples are synthesized on Oxford Flower-17 dataset using IGSM.\fi}
	\resizebox{\linewidth}{!}{
		\begin{tabular}{l rrrr}
			\toprule
			\textbf{Kernel Sizes} & $7\times7$ & $11\times11$ & $15\times15$ & $21\times21$ \\
			\midrule
			\textbf{Top-1 Accuracy} & 85.29\% & 86.18\% & \textbf{87.35\%} & 86.18\% \\
			\bottomrule
		\end{tabular}
	}
	\label{table:ablation_on_kernel_size}
\end{table}

We evaluate four different kernel sizes of the first convolution layer, with Image Steps set to 20 and Network Steps set to 300. TABLE \ref{table:ablation_on_kernel_size} shows that our method performs the best with kernel size $15\times15$. Small receptive field is susceptible to adversarial perturbations. Convolutions with smaller receptive field benefit the reconstruction of image details, even including the adversarial perturbation on the given inference image. On the contrary, images synthesized by convolutions with larger receptive field may lower the quality of small patterns and details, which also degrades the classification accuracy.

\subsection{Investigation with Natural/Clean Images}
We present the results on both clean and adversarial samples, as shown in TABLE~\ref{table:investigation_with_clean}. The results are obtained on Oxford Flower-17 with C\&W as attack and ResNet18 as target classifier. ‘None’ denotes the target classifier without any defenses. In TABLE~\ref{table:investigation_with_clean}, JPEG and Pixel Defend are the best on clean images, but their top-1 accuracy are less than 40\% and the worst on adversarial images. TVM is worse than our method on both clean and adversarial samples. Pixel Deflection is close to our method on clean images but our method exceeds it by 5.59\% top-1 accuracy against the adversarial attack. Randomization surpasses our method by 2.9\% on clean images but our method outperforms it by 8.6\% on adversarial examples. {\color{black}Comparing with DIP, our proposed method obtains 1.5\% higher top1-accuracy against adversarial samples, and 2.4\% lower accuracy on clean images. Overall, the proposed method achieves the state-of-the-art trade-off between natural images and adversarial examples.}
\begin{table}[!t]
	\centering
	\small
	\caption{\color{black}Investigation with natural/clean images.}
	\begin{tabular}{l rr}
		\toprule
		\textbf{Defense Methods} & \textbf{Clean Images} & \textbf{Adversarial} \\
		\midrule
		None             & 97.06\%        & 0.00\%  \\
		Mean Filter~\cite{li2017adversarial}   & 94.71\%        & 66.47\% \\
		JPEG~\cite{Dziugaite2016A}             & 97.06\%        & 28.82\% \\
		TVM~\cite{guo2018countering}           & 90.88\%        & 80.29\% \\
		Pixel Deflection~\cite{prakash2018deflecting} & 92.35\%        & 78.82\% \\
		Randomization~\cite{xie2018mitigating}    & 95.00\%        & 75.88\% \\
		Pixel Defend~\cite{song2018pixeldefend}     & 96.76\%        & 37.35\% \\
		DIP~\cite{kattamis2019exploring}       & 94.41\%        & 82.94\% \\
		Ours             & 92.06\%        & 84.41\% \\
		\bottomrule
	\end{tabular}
	\label{table:investigation_with_clean}
\end{table}
\subsection{Visualization of Online Alternate Generation}
This section visualizes the synthesized image of our defense method during the alternate generation. As shown in Fig.~\ref{fig:ILSVRC_Flower}, the columns captioned with `Iteration $k$' are the intermediate synthesized images of our proposed defense method at iteration $k$. The maximum number of iterations corresponds to \textit{Network Steps} that controls how many times the neural model in our method is updated. As the iteration number increases, the synthesized image become clearer and sharper. The synthesized image at the last iteration is visually similar to the original image. The column captioned with (a) is the \textit{residual map} of an adversarial example (synthesized by Iterative FGSM). The residual map of a synthesized image is defined as a normalized pixel-wise absolute difference between the synthesized image and its corresponding original image. The column captioned with (b) is the residual map of the synthesized image in our method. Comparing column (a) with column (b), the residual map of an adversarial examples differs from that of the synthesized image in our method, which suggests that the noise distribution between an adversarial example and its corresponding synthesized image is quite distinct. The synthesized images could be less affected by the adversarial noises. 

\subsection{Whether Alternate Update is Necessary}
To understand whether the alternate update scheme in our proposed method is necessary, let us consider a simple auto-encoder. The auto-encoder employs online training strategy as the proposed method to achieve portable defense, but does not adopt two-step update. Given an image, we first randomly initialize the parameters of the auto-encoder, and then online tune its parameters by taking the image as input and supervision. After $T_N$-step updates, the auto-encoder takes the image as input and outputs another image. The output serves as a substitute to be sent into a target classifier. For fair comparison, we instantiate the auto-encoder in a symmetric form, and its encoder part as the same as the proposed generator. The auto-encoder consists of a convolution layer, two fully-connected layers, a non-linear activation and a deconvolution layer. The fully-connected layers are in between the convolution and the deconvolution while the non-linear layer is located in between these two fully-connected layers. The first fully-connected layer encodes a tensor into a scalar while the second one decodes a scalar into a tensor. The auto-encoder achieves 82.94\% on Oxford Flower-17 benchmark with C\&W as attack and ResNet18 as target classifier. Our proposed method using alternate update obtains 84.41\%, slightly better than the auto-encoder using one-step update. It may be due to that the alternate update with Langevin Dynamics could better sample a representative point in some latent space, and hence recover more accurate semantics for an input image.

\section{Conclusion}
In this paper we develop a novel portable defense framework that reconstructs an image with less adversarial noise and almost the same semantics as an input image. The reconstructed image acts as a substitute to defend against adversarial attacks. 
The hyper-parameters of our proposed defense do not need to be tuned on any adversarial examples, which avoid a bias towards some known attacks. The proposed defense does not access, modify any parameters or intermediate outputs of target models, which allows the defense portably transferable to a wide range of target classifiers.
Experimental results show that our method obtains the state-of-the-art performance.

\ifCLASSOPTIONcaptionsoff
  \newpage
\fi



%



\bibliographystyle{IEEEtran}
\bibliography{paper}

%

\begin{IEEEbiography}[{\includegraphics[width=1in,height=1.25in,clip,keepaspectratio]{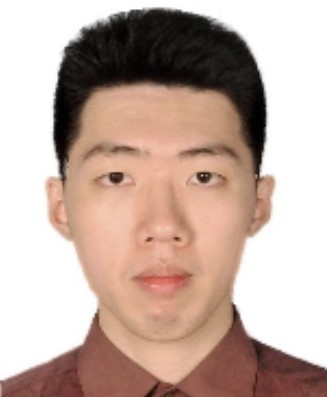}}]{Haofeng Li}
is a research scientist in Shenzhen Research Institute of Big Data, The Chinese University of Hong Kong (Shenzhen). He received his Ph.D. degree from the Department of Computer Science, the University of Hong Kong in 2020, and his B.Sc. degree from School of Data and Computer Science, Sun Yat-Sen University in 2015. He is a recipient of Hong Kong PhD Fellowship. He has been serving as a reviewer for TIP, Pattern Recognition, Neurocomputing and IEEE Access. His current research interests include computer vision, image processing and deep learning.
\end{IEEEbiography}

\begin{IEEEbiography}[{\includegraphics[width=1in,height=1.25in,clip,keepaspectratio]{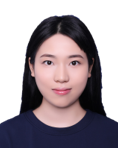}}]{Yirui Zeng}
received her B.S. degree in the School of Electronics and Information engineering , Chongqing University in 2017. She is currently pursuing the master’s degree in the School of Electronics and Information engineering, Sun Yat-Sen University. Her current research interests include computer vision and deep learning.
\end{IEEEbiography}


\begin{IEEEbiography}[{\includegraphics[width=1in,height=1.25in,clip,keepaspectratio]{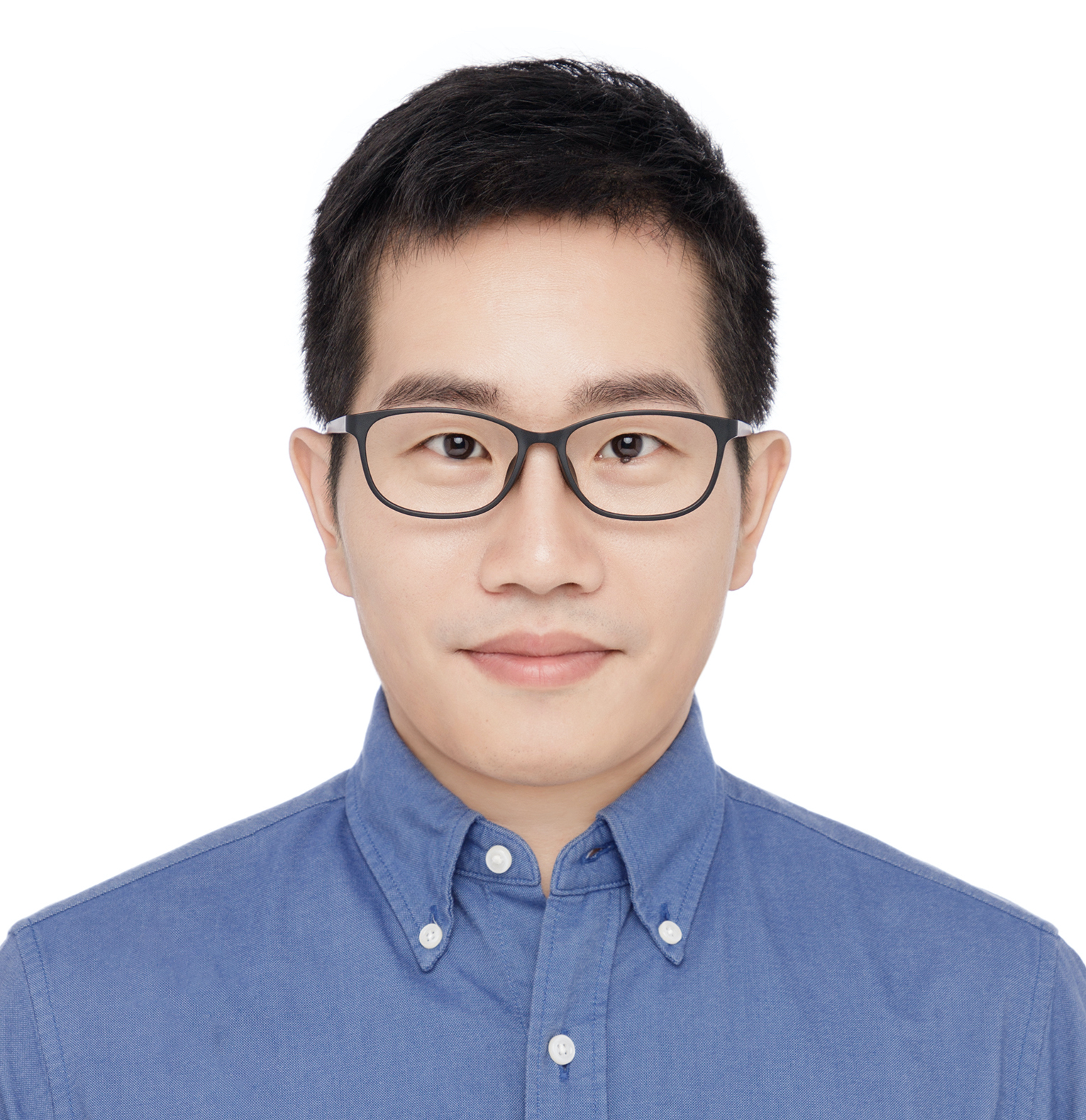}}]{Guanbin Li} (M'15) is currently an associate professor in School of Data and Computer Science, Sun Yat-sen University. He received his PhD degree from the University of Hong Kong in 2016. His current research interests include computer vision, image processing, and deep learning. He is a recipient of ICCV 2019 Best Paper Nomination Award. He has authorized and co-authorized on more than 60 papers in top-tier academic journals and conferences. He serves as an area chair for the conference of VISAPP. He has been serving as a reviewer for numerous academic journals and conferences such as TPAMI, IJCV, TIP, TMM, TCyb, CVPR, ICCV, ECCV and NeurIPS.
\end{IEEEbiography}

\begin{IEEEbiography}[{\includegraphics[width=1in,height=1.25in,clip,keepaspectratio]{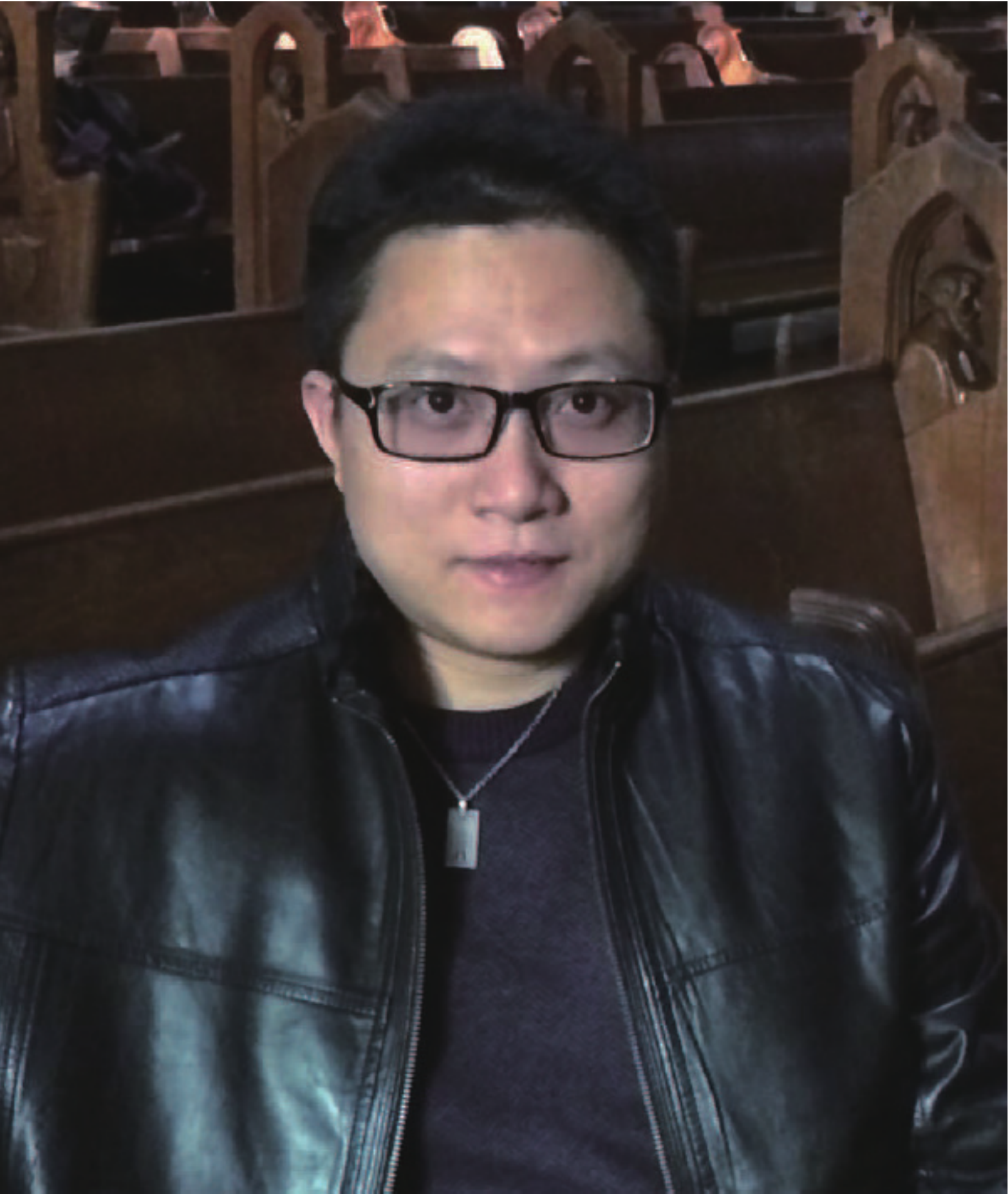}}]{Liang Lin}
(M'09, SM'15) is a full Professor of Sun Yat-sen University. He is the Excellent Young Scientist of the National Natural Science Foundation of China. From 2008 to 2010, he was a Post-Doctoral Fellow at the University of California, Los Angeles. From 2014 to 2015, as a senior visiting scholar, he was with the Hong Kong Polytechnic University and the Chinese University of Hong Kong. He currently leads the SenseTime R$\&$D teams to develop cutting-edge and deliverable solutions on computer vision, data analysis and mining, and intelligent robotic systems. He has authored and co-authored more than 100 papers in top-tier academic journals and conferences. He has been serving as an associate editor of IEEE Trans. Human-Machine Systems, The Visual Computer and Neurocomputing. He served as Area/Session Chair for numerous conferences, including ICME, ACCV, and ICMR. He was the recipient of the Best Paper Runners-Up Award at ACM NPAR 2010, a Google Faculty Award in 2012, the Best Paper Diamond Award at IEEE ICME 2017, and the Hong Kong Scholars Award in 2014. He is a Fellow of IET.
\end{IEEEbiography}

\begin{IEEEbiography}[{\includegraphics[width=1in,height=1.25in,clip,keepaspectratio]{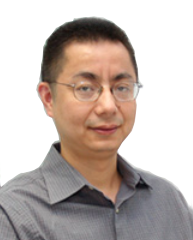}}]{Yizhou Yu} (M'10, SM'12, F'19) received the PhD degree from University of California at Berkeley in 2000. He is a professor at The University of Hong Kong, and was a faculty member at University of Illinois at Urbana-Champaign for twelve years. He is a recipient of 2002 US National Science Foundation CAREER Award, 2007 NNSF China Overseas Distinguished Young Investigator Award, and ACCV 2018 Best Application Paper Award. Prof Yu has served on the editorial board of IET Computer Vision, The Visual Computer, and IEEE Transactions on Visualization and Computer Graphics. He has also served on the program committee of many leading international conferences, including SIGGRAPH, SIGGRAPH Asia, and International Conference on Computer Vision. His current research interests include computer vision, deep learning, biomedical data analysis, computational visual media and geometric computing.
\end{IEEEbiography}




\end{document}